\documentclass[12pt]{article}

\usepackage[utf8]{inputenc}
\usepackage{graphicx}
\usepackage{subcaption}
\usepackage{verbatim}
\usepackage{enumitem}
\usepackage{csquotes}
\usepackage{amsmath,amssymb}
\usepackage{listings} 
\usepackage{algorithm}
\usepackage{algorithmic}
\usepackage{CJKutf8}
\usepackage{float} 
\usepackage{tikz}
\usepackage{pgfplots}
\usepackage[colorlinks=true,linkcolor=blue,citecolor=blue]{hyperref}
\usepackage[left=1in,right=1in,top=1in,bottom=1in]{geometry}

\usetikzlibrary{shapes.geometric}
\usetikzlibrary{arrows}
\usetikzlibrary{shapes.arrows}
\usetikzlibrary{positioning}

\let\origfigure\figure
\let\endorigfigure\endfigure
\renewenvironment{figure}[1][2] {
    \expandafter\origfigure\expandafter[H]
} {
    \endorigfigure
}

\pgfplotsset{compat=1.16}

\title{Recommending Actionable Strategies: A Semantic Approach to Integrating Analytical Frameworks with Decision Heuristics}

\author{
    Renato Ghisellini$^{1}$, 
    Remo Pareschi$^{2}$\thanks{Corresponding author: remo.pareschi@unimol.it. ORCID: 0000-0002-4912-582X}, 
    Marco Pedroni$^{1}$, 
    Giovanni Battista Raggi$^{1}$\\
    \small{$^{1}$[Institute for Generative Strategy]}\\
    \small{$^{2}$[Stake Lab, University of Molise]}
}

\date{\today}

\begin{document}
\maketitle

\begin{abstract}
We present a novel approach for \textbf{recommending actionable strategies} by integrating strategic frameworks with decision heuristics through \textbf{semantic analysis}. While strategy frameworks provide systematic models for assessment and planning, and decision heuristics encode experiential knowledge, these traditions have historically remained separate. Our methodology bridges this gap using \textbf{advanced natural language processing (NLP)}, demonstrated through integrating frameworks like the 6C model with the Thirty-Six Stratagems. The approach employs \textbf{vector space representations} and \textbf{semantic similarity calculations} to map framework parameters to heuristic patterns, supported by a computational architecture that combines deep semantic processing with constrained use of Large Language Models. By processing both \textbf{primary content} and \textbf{secondary elements} (diagrams, matrices) as complementary linguistic representations, we demonstrate effectiveness through corporate strategy case studies. The methodology \textbf{generalizes} to various analytical frameworks and heuristic sets, culminating in a \textbf{plug-and-play architecture} for generating \textbf{recommender systems} that enable cohesive integration of strategic frameworks and decision heuristics into actionable guidance.
\end{abstract}

\section{Introduction}

Organizations today rely on two primary but historically separate textual traditions for \textbf{strategic management and decision-making}: \textbf{analytical frameworks} and \textbf{decision heuristics}. Both traditions encode strategic knowledge in natural language—often complemented by diagrams or structured representations—yet differ in scope and style. Analytical frameworks such as Porter’s Five Forces, SWOT Analysis, and Value Chain Analysis offer systematic lenses for situational assessment and long-term planning. Meanwhile, decision heuristics—from early military wisdom (e.g., the Thirty-Six Stratagems) to modern “rules of thumb”—provide concise, actionable insights forged through real-world experience.

In practice, \textbf{combining these two traditions} holds tangible advantages: a more balanced approach to strategic planning, clearer avenues for \emph{evidence-based recommendations}, and reduced time spent on exhaustive analyses. However, frameworks and heuristics rarely interact in a unified process. Frameworks excel at comprehensiveness and rigor but risk \emph{analysis paralysis}, whereas heuristics are more agile but \emph{can oversimplify} complex scenarios. \textbf{Bridging this gap} would enable decision-makers to reap the complementary strengths of each method, yielding recommended strategies that are both thorough and swiftly implementable.

Recent advances in \textbf{artificial intelligence (AI) and natural language processing (NLP)} offer a powerful way to integrate these traditions. By applying \textbf{semantic analysis} to uncover linguistic patterns, rhetorical structures, and conceptual interdependencies within strategic texts, we can construct automated mappings between the methodical constructs of analytical frameworks and the concise action steps of heuristics. This paper proposes a \textbf{recommender-system architecture} that leverages these mappings to generate \textbf{actionable strategic recommendations}, ultimately expediting decision-making and enhancing strategic insight.

To illustrate this \textbf{semantic integration} concretely, we focus on two representative models: the 6C framework and the Thirty-Six Stratagems. The 6C framework synthesizes recurring strategic themes (offensive/defensive strength, relational capacity, potential energy, temporal availability, and context-fit), drawn from military and business literature. In contrast, the Thirty-Six Stratagems—rooted in Chinese political, military, and civil discourse—encapsulate centuries of heuristic insight in pithy expressions. Through \textbf{advanced NLP} tools—such as vector-space embeddings, topic modeling, and pattern recognition—we demonstrate how linguistic cues in each stratagem correlate with specific 6C parameters. This systematic analysis then drives an automated pipeline that matches any given strategic situation with suitable heuristics, producing \textbf{evidence-based, context-aware recommendations}.

Two key innovations underscore our approach. First, we embed the system in an \textbf{interactive simulation environment}, prompting decision-makers to express scenarios in natural language. The environment analyzes these textual inputs, computes relevance scores, and returns \textbf{recommendations} for how best to combine or select heuristics in light of the chosen strategic framework. Second, we employ \textbf{Large Language Models (LLMs)} in a controlled manner to produce coherent, narrative-style reports that clarify the rationale behind each recommendation. By integrating LLMs as interpreters \emph{rather} than autonomous decision-makers, we preserve analytical rigor while providing accessible explanations.

In what follows, we detail our \textbf{semantic methodology}, discuss the \textbf{computational architecture} that enables framework-heuristic integration, and illustrate its real-world relevance through \textbf{case studies} in corporate strategy. We then show how this \textbf{plug-and-play architecture} generalizes beyond 6C and the Thirty-Six Stratagems, adapting to other widely known frameworks such as Porter’s Five Forces and SWOT. Ultimately, we aim to demonstrate how \textbf{organizations can deploy a recommender-system approach} to merge comprehensive strategic analysis with proven heuristic insights, delivering \textbf{actionable guidance} that is both robust and readily applicable in complex environments.

The remainder of this paper is structured as follows:
\vspace{1em}
\begin{itemize}[itemsep=2pt,topsep=2pt]
\item Section \ref{background} provides background knowledge
    \item Section \ref{language-analysis} details our language analysis methodology for framework integration
    \item Section \ref{architecture} presents the computational architecture supporting this integration
    \item Section \ref{casestudies} demonstrates the approach through two case studies
    \item Section \ref{empirical} provides empirical validation
    \item Section \ref{related} discusses related work
    \item Section \ref{conclusion} discusses implications and future directions
\end{itemize}

\section{Background}
\label{background}

Integrating analytical frameworks with decision heuristics through semantic analysis represents a highly interdisciplinary endeavor, drawing from multiple domains, including strategic management, heuristics, computer science, and linguistics. This confluence of fields necessitates thoroughly examining key concepts and prior work across several dimensions. Specifically, we must understand (1) how strategic frameworks systematize decision parameters, as exemplified by the 6C model; (2) how decision heuristics encapsulate experiential knowledge, illustrated through the Thirty-Six Stratagems; (3) how semantic analysis enables framework-heuristic integration; (4) how mathematical formulations like Kullback-Leibler divergence support validation; and (5) how gamification principles and Large Language Models facilitate practical implementation. The following subsections provide this essential foundation, which underlies our novel recommender-system architecture for strategic decision support.

\subsection{The 6C Framework as an Analytical Classification Tool}
\label{background-6c}

The 6C framework was conceived to provide a set of clear, well-defined parameters that would enable straightforward experimentation with our semantic integration approach. While established frameworks like SWOT Analysis and Porter's Five Forces offer comprehensive analytical tools, the 6C parameters were specifically designed to facilitate initial testing and validation of our methodology, with the understanding that the same principles could then be transferred to these well-known and widely adopted frameworks. The parameters were distilled from an extensive study of strategic literature, offering a simplified yet robust foundation for our initial framework-heuristic integration experiments.

\vspace{1em}
The key parameters of the 6C model are as follows:
\begin{enumerate}[itemsep=2pt,topsep=2pt]
    \item \textbf{$p_1$ = Offensive Strength:} The ability to proactively shape and influence the strategic landscape.
    \item \textbf{$p_2$ = Defensive Strength:} The resilience to respond effectively to adversarial actions or challenges.
    \item \textbf{$p_3$ = Relational Capacity:} The ability to manage and leverage relationships with external stakeholders.
    \item \textbf{$p_4$ = Potential Energy:} The availability and strategic deployment of resources.
    \item \textbf{$p_5$ = Temporal Availability:} The strategic use of time and timing in decision-making.
    \item \textbf{$p_6$ = Contextual Fit:} The degree to which decisions align with the strategic context, ensuring they are well-informed and relevant.
\end{enumerate}

These classification processes emerged from a comparative study of prominent military strategists---ranging from ancient figures such as Sun Tzu and Chanakya to modern thinkers like Machiavelli and Clausewitz and more contemporary theorists such as Beaufre and Liddell Hart. Over time, they have transcended their military origins to influence corporate strategy, echoing broader perspectives that integrate historical wisdom with rigorous managerial concepts \cite{Mintzberg2005StrategySafari}.

\vspace{1em}
The 6C framework functions as a classification system for organizing and analyzing data obtained through competitive intelligence and data analytics, a capability especially valuable in digitally transforming industries \cite{hirt2014digital}. This classification provides a structured approach to understanding competitive landscapes, where the six parameters can be assessed quantitatively or qualitatively. While we do not delve into the specific mechanics of setting each parameter---this is handled through the gamified environment described in Section~\ref{background-gamification}---the focus of this paper is on \emph{how} these parameters (once established) can be integrated with heuristic decision patterns via semantic analysis.

\subsection{The Thirty-Six Stratagems: Crystallized Decision Patterns}
\label{background-36}

The Thirty-Six Stratagems comprise a collection of ancient Chinese military decision heuristics that have evolved into widely applicable strategic principles \cite{taylor2013thirty,von1991book}. These stratagems are especially valuable for our study since they embody concise, experience-based rules, akin to "simple rules" in strategy-making \cite{Eisenhardt2001SimpleRules}, but encoded in metaphorical language with deep historical roots. The highly metaphorical nature of these stratagems, while encoding deep strategic wisdom, presents unique challenges for semantic analysis. Their idiosyncratic expressions require careful interpretation to bridge ancient military metaphors with modern strategic concepts. This linguistic complexity makes them an especially rigorous test case for our framework integration methodology---success here would suggest strong generalizability to more straightforward decision-making frameworks.

\vspace{1em}
Traditionally, the thirty-six stratagems are divided into six categories:
\begin{enumerate}[itemsep=2pt,topsep=2pt]
    \item Stratagems for winning advantageous positions
    \item Stratagems for confrontation
    \item Stratagems for attack
    \item Stratagems for creating confusion
    \item Stratagems for gaining ground
    \item Stratagems for desperate situations
\end{enumerate}

\vspace{1em}
% Chinese characters in the following maxims
Each stratagem defines a concrete pattern of strategic action---often encapsulated in pithy maxims such as ``Besiege Wei to rescue Zhao'' (\begin{CJK*}{UTF8}{gbsn}围魏救赵\end{CJK*}), which illustrates an indirect approach, or ``Kill with a borrowed knife'' (\begin{CJK*}{UTF8}{gbsn}借刀杀人\end{CJK*}), which suggests leveraging external resources. By codifying these patterns, the Thirty-Six Stratagems lend themselves to computational extraction and matching when we embed them in modern NLP pipelines.

\subsection{Additional Strategic Frameworks}
\label{background-frameworks}

While the 6C framework and the Thirty-Six Stratagems serve as primary examples in our study, our approach generalizes to multiple analytical frameworks. Following the systematic nature of the 6C framework, other prominent analytical frameworks provide structured parameters for strategic assessment:

\vspace{1em}
\paragraph{SWOT Analysis}
SWOT Analysis, developed by Albert Humphrey at Stanford Research Institute, offers a systematic approach to evaluating internal and external factors \cite{helms2010exploring}. Its clear parameter structure makes it particularly suitable for semantic analysis:
\begin{itemize}[itemsep=2pt,topsep=2pt]
\item \textbf{Strengths:} Internal capabilities and resources that provide competitive advantages
\item \textbf{Weaknesses:} Internal limitations that may hinder strategic objectives
\item \textbf{Opportunities:} External factors or trends that could benefit the organization
\item \textbf{Threats:} External challenges that could negatively impact performance
\end{itemize}

\vspace{1em}
\paragraph{Porter's Five Forces}
This analytical framework, developed by Michael Porter, systematically dissects industry structure through five well-defined parameters \cite{porter2008five}:
\begin{itemize}[itemsep=2pt,topsep=2pt]
\item \textbf{Competitive Rivalry:} Intensity of competition among existing players
\item \textbf{Supplier Power:} Bargaining power of suppliers
\item \textbf{Buyer Power:} Bargaining power of customers
\item \textbf{Threat of New Entrants:} Ease with which new competitors can enter
\item \textbf{Threat of Substitution:} Availability of alternative products or services
\end{itemize}

\vspace{1em}
Like the 6C framework, SWOT and Five Forces provide clear analytical parameters that can be vectorized and processed through semantic analysis for framework-heuristic integration. The effectiveness of our approach with these widely adopted frameworks demonstrates its potential applicability to other strategic analysis tools, both contemporary and classical, a direction we discuss in our future work.

\subsection{Semantic Analysis in Strategic Text Processing}
\label{background-semantic}

Our approach employs several key techniques from Natural Language Processing (NLP) to analyze and connect strategic frameworks (like the 6C model) with decision heuristics (such as the Thirty-Six Stratagems). Recent advances in NLP, especially \emph{Transformer}-based architectures~\cite{vaswani}, have shown that vector-space language representations can capture nuanced semantic meaning and contextual relationships.

\vspace{1em}
\begin{itemize}[itemsep=2pt,topsep=2pt]
    \item \textbf{Vector Space Representations:} We encode strategic concepts using word embeddings and sentence transformers, building on methods such as \emph{BERT}~\cite{devlin}, \emph{Sentence-BERT}~\cite{reimers}, that excel in capturing contextual nuances. These techniques enable mathematical operations on semantic meaning and allow us to compare entire passages or phrases in a high-dimensional embedding space.
    
    \item \textbf{Topic Modeling (Optional)}: We can also apply \emph{Latent Dirichlet Allocation} (LDA) and related approaches~\cite{blei2012probabilistic} to identify high-level themes in strategic texts (e.g., "alliances," "resource optimization"). Although not the core driver in our current implementation, such thematic analysis can support explainability by highlighting relevant topics that link to each framework parameter or stratagem.
    
    \item \textbf{Semantic Similarity Metrics:} Using cosine similarity or similar measures, we quantify relationships between vectors representing framework parameters and stratagem patterns. In our implementation with the Thirty-Six Stratagems, this objectively measures how well each stratagem aligns with specific 6C parameters (e.g., \emph{Offensive Strength, Relational Capacity}).
\end{itemize}

\vspace{1em}
The semantic analysis process involves the following steps:
\begin{enumerate}[itemsep=2pt,topsep=2pt]
    \item Preprocessing (if needed) and Concept Extraction: 
    In scenarios where texts are unstructured, we may apply standard NLP techniques 
    (tokenization, chunking, domain ontology extraction). However, this step becomes straightforward if frameworks 
    like 6C or Porter's Five Forces are already delineated.
    \item Creating vector representations of framework parameters and heuristic patterns;
    \item Computing similarity matrices to link frameworks (e.g., 6C) with heuristics (e.g., Thirty-Six Stratagems);
    \item Identifying significant semantic connections and ranking them for further interpretation.
\end{enumerate}

\vspace{1em}
Our work builds on emerging applications of BERT-like methods to nontraditional NLP contexts \cite{pareschiListening}, where domain-specific texts require fine-grained semantic understanding. Adapting these cutting-edge techniques preserves both the textual richness of each strategic expression and the interpretability needed for real-world strategic decision-making. 

Having said that, it must be added that although these models have proven effective, they inherit statistical biases from their training corpora, which may affect how parameters and heuristics are aligned. Because we employ LLMs only for explanatory output, the overall recommendation pipeline remains largely heuristic- and framework-driven. Nevertheless, mitigating potential embedding biases---by fine-tuning domain-specific corpora, employing bias detection tools, or adopting interpretability frameworks---is an important direction for future development.

While metrics like the symmetric Jensen--Shannon divergence could be considered, KL divergence provides an efficient and intuitive tool for our semantic analysis workflow, where we compare system-discovered parameter distributions to expert-annotated ones. This approach enables quantitative validation of our semantic mappings.

\vspace{1em}
\subsection{Kullback--Leibler Divergence}
\label{background-kl}

In information theory and statistics, the \textbf{Kullback--Leibler (KL) divergence}
measures how one probability distribution diverges from a second,
\emph{reference} distribution~\cite{CoverThomas2006}. Given two discrete
probability distributions (P) and (Q) over the same outcome space
($\Omega$), the KL divergence is defined as:

\begin{equation}
\label{eq:kl-divergence-background}
D_{\mathrm{KL}}(P \,|\, Q) = \sum_{i \in \Omega}
P(i)\log\Bigl(\frac{P(i)}{Q(i)}\Bigr).
\end{equation}

Intuitively, if (P) represents the ``true'' or expert-labeled distribution
(e.g., the relative importance of each parameter in a scenario), while
(Q) is the model's approximate distribution, the KL divergence quantifies how
inefficient it is to use (Q) in place of (P). A lower $D_{\mathrm{KL}}(P \,|\, Q)$
indicates a closer match between the two distributions, while higher values
signify greater disparity. Unlike many distance metrics, KL divergence is
\emph{not} symmetric (i.e., $D_{\mathrm{KL}}(P \,|\, Q) \neq
D_{\mathrm{KL}}(Q \,|\, P)$) which means the direction of comparison matters. Because KL divergence also fails to satisfy the triangle inequality, it is not a true metric in the formal sense; however, it remains a widely used 'distance-like' measure for comparing probability distributions.

\vspace{1em}
\paragraph{\emph{Why KL Divergence?}}
We selected KL divergence for three key reasons:
\begin{itemize}[itemsep=2pt,topsep=2pt]
\item \textbf{Interpretability:} It offers a straightforward interpretation of the ``cost'' of using an approximate distribution, aligning well with our need to validate semantic analysis against expert annotations.
\item \textbf{Directionality:} Its asymmetric nature suits our context, where we specifically care about how well our approximate distributions match expert distributions, rather than vice versa.
\item \textbf{Established Usage:} Its widespread adoption in machine learning and information theory provides a well-tested foundation for measuring distributional differences.
\end{itemize}

\vspace{1em}
While metrics like the symmetric Jensen--Shannon divergence could be considered, KL divergence provides an efficient and intuitive tool for our semantic analysis workflow, where we compare system-discovered parameter distributions to expert-annotated ones. A lower divergence indicates our system captures expert priorities effectively, while higher values highlight areas needing further calibration.

\vspace{1em}

\paragraph{\emph{Usage in This Study}}
In our context, we leverage KL divergence to compare the \emph{discovered} parameter distributions 
(e.g., derived from the system's semantic analysis) to \emph{expert-annotated} distributions. This 
comparison provides a quantitative measure of how closely our system's interpretations align with 
domain experts' judgments, thereby helping to validate and refine the \emph{robustness} of our 
semantic mapping process. A lower divergence indicates our system captures expert priorities 
effectively, while higher values highlight areas needing further calibration.

\subsection{Gamification of Strategic Decision-Making}
\label{background-gamification}

To streamline the usage of these analytic results, we have developed a prototype \emph{interactive simulation environment} that enables:

\begin{itemize}[itemsep=2pt,topsep=2pt]
    \item Exploration of different strategic scenarios;
    \item Testing of various decision combinations;
    \item Immediate feedback on potential outcomes;
    \item Gradual learning from simulated experiences.
\end{itemize}

\vspace{1em}
Such \emph{gamification} introduces a user-centric interface through which decision-makers set or adjust the situation-specific scores of the chosen analytical framework. For instance, users can enter or modify the 6C parameters---\emph{Offensive Strength}, \emph{Defensive Strength}, and so on---based on real competitive intelligence data or hypothetical "what-if" explorations. The system then applies its semantic mappings (Section~\ref{background-6c}) to generate immediate feedback on how each parameter configuration impacts recommended strategies or outcomes. 

\vspace{1em}
Although this gamified environment is fully implemented and was used to produce the case studies described in this paper, its detailed interface design and mechanics lie beyond our current scope. Such a \emph{user-interface} analysis merits its own dedicated treatment and can be assumed here without loss of relevant information about the underlying framework. By highlighting interactive experimentation and rapid feedback, our approach resonates with broader digital transformation trends~\cite{hirt2014digital} and encourages deeper engagement and practical experimentation among corporate decision-makers.

\section{Language Analysis Methodology}
\label{language-analysis}

Our methodology for integrating analytical frameworks with decision heuristics centers on semantic analysis of strategic texts. This section details the technical approach to discovering and quantifying relationships between framework parameters and heuristic patterns.

\subsection{Vector Space Representation}
\label{vector-space}

The first step involves creating vector representations of both framework parameters and heuristic descriptions:
\begin{equation}
\label{eq:vector-representation}
v(t) = \sum_{w \in t} \alpha_w \cdot e(w)
\end{equation}
where $v(t)$ is the vector representation of text $t$, $w$ represents individual words or phrases, $e(w)$ is the embedding vector for word $w$, and $\alpha_w$ is the weight assigned to word $w$. For framework parameters (in our case, the 6Cs), we create vectors from their definitions and associated descriptive text:
\begin{equation}
\label{eq:parameter-vector}
p_i = v(d_i) + \lambda \sum_{j} v(c_{ij})
\end{equation}
where $p_i$ is the vector for parameter $i$, $d_i$ is the base definition of parameter $i$, $c_{ij}$ are associated contextual descriptions, and $\lambda$ is a weighting factor for contextual information. Detailed calculations and examples of this vector space representation are provided in Appendix \ref{appendix:vector-space}.

\subsection{Semantic Similarity Computation}
\label{semantic-similarity}

We compute semantic similarity between framework parameters and heuristics using the cosine similarity measure. Specifically, for a given parameter vector $p_i$ and heuristic vector $h_j$, the similarity is:
\begin{equation}
\label{eq:semantic-similarity}
\text{sim}(p_i, h_j) = \frac{p_i \cdot h_j}{\|p_i\|\|h_j\|}
\end{equation}

\noindent
where $p_i \cdot h_j$ denotes the dot (scalar) product of the two vectors,
\[
p_i \cdot h_j = \sum_{k} (p_i)_k (h_j)_k,
\]
and $\|p_i\|$ (likewise $\|h_j\|$) is the Euclidean norm,
\[
\|p_i\| = \sqrt{\sum_{k}((p_i)_k)^2}.
\]

\vspace{1em}
A higher cosine similarity value $\text{sim}(p_i, h_j)$ indicates a closer semantic relationship between parameter $p_i$ and heuristic $h_j$.

\vspace{1em}
Computing $\text{sim}(p_i, h_j)$ for all parameters $i \in \{1, \ldots, m\}$ and all heuristics $j \in \{1, \ldots, n\}$ produces a similarity matrix $S$, in which each element $s_{ij}$ is defined by:
\begin{equation}
\label{eq:similarity-matrix}
S = \begin{bmatrix}
   s_{11} & s_{12} & \cdots & s_{1n} \\
   s_{21} & s_{22} & \cdots & s_{2n} \\
   \vdots & \vdots & \ddots & \vdots \\
   s_{m1} & s_{m2} & \cdots & s_{mn}
\end{bmatrix}
\quad\text{with}\quad
s_{ij} = \text{sim}(p_i, h_j).
\end{equation}

For example, when comparing parameter $p_3$ (Relational Capacity) with Stratagem~24 (\enquote{Use Allies' Resources}), our system computes a similarity score of $0.93$, quantitatively capturing their strong semantic alignment. Detailed calculations and further examples of this process appear in Appendix~\ref{appendix:similarity-calc}.

\subsection{Distribution Discovery}
\label{distribution-discovery}

For each heuristic $h_j$, our system generates a \emph{discovered distribution} across the framework parameters (e.g., the 6C parameters). Concretely, we first compute similarity scores $s_{ij}$ between parameter $p_i$ and heuristic $h_j$ (Section~\ref{semantic-similarity}). We then normalize these scores to form a probability-like distribution:
\begin{equation}
\label{eq:parameter-distribution}
d_{ij} = \frac{s_{ij}}{\sum_{k} s_{kj}},
\end{equation}
where $d_{ij}$ represents the weight (or relative importance) of parameter $p_i$ in heuristic $h_j$. This summation-based (L1) normalization treats each heuristic's parameter weights as if they were probabilities that sum to 1. It thus naturally encodes the idea that a given heuristic distributes its ``attention'' across the available parameters.

\vspace{1em}
\noindent
\textbf{Note on Alternative Normalization.}\, 
An alternative would be to use the Euclidean (L2) norm, where
\[
d_{ij} = \frac{s_{ij}}{\sqrt{\sum_{k} (s_{kj})^2}},
\]
thereby turning each heuristic's parameter vector into a \emph{unit vector} in $\ell_2$ space. In our approach, we opt for L1 normalization to mirror a "probability-like" interpretation---each heuristic can be seen as distributing its "weight" over parameters in a manner analogous to probabilities. L2 normalization could be equally valid in other contexts, especially if one prefers strictly geometric interpretations of distance in the parameter space.

\vspace{1em}
To validate these distributions, we compare them against expert-annotated distributions using the Kullback-Leibler (KL) divergence measure. For Stratagem 24, this comparison yielded a KL divergence of 0.0273, indicating strong alignment between system-discovered and expert-provided distributions. Detailed calculations and additional examples of this validation process are provided in Appendix~\ref{appendix:kl-calc}.

\vspace{1em}
By comparing discovered distributions against expert-annotated ones for each heuristic, we obtain a numerical sense of alignment or mismatch. This process can be iterated: a large KL divergence flags a heuristic whose vector representation needs either textual refinements (e.g., additional synonyms or clarifications) or updates to the weighting scheme. Over time, \textbf{machine-based distribution discovery} converges with expert insights, yielding robust mappings that faithfully reflect how these strategic heuristics fit into analytical parameters.

\vspace{1em}
As we already pointed out, the \emph{experts} in this step are specialists in the \emph{methodologies} (e.g., the 6C model, the Thirty-Six Stratagems) rather than sector-specific domain experts. This distinction ensures that the high-level semantic structure of each stratagem is validated by those who understand it conceptually, independent of particular industries or case studies.
\subsection{Stratagem Selection Algorithm}
\label{stratagem-selection}

After the \emph{distribution discovery} phase (Section~\ref{distribution-discovery}), we obtain an \textbf{invariant} distribution of analytical properties for each heuristic. This invariant distribution reflects how strongly each heuristic (e.g., a particular stratagem) aligns with each framework parameter (e.g., the 6C model) based on the text analysis and expert validation.

\vspace{1em}
\noindent
\textbf{Situation-Specific Parameter Vector.}\,
In contrast, a \textbf{variable} (or \emph{current}) situation vector, denoted by $x$, describes how the analytical parameters apply to the \emph{present scenario}. For instance, if a certain strategic context demands high \emph{Offensive Strength} ($p_1$) and moderate \emph{Relational Capacity} ($p_3$), $x$ will capture these intensities accordingly. 

\vspace{1em}
\noindent
\textbf{Matching Heuristics to the Situation.}\,
By comparing $x$ with each heuristic's invariant distribution, we produce a \emph{recommendation score} that indicates how well that heuristic fits the present conditions. This process is summarized in Algorithm~\ref{alg:stratagem-selection}. 

\begin{algorithm}[H]
\caption{Stratagem Selection}
\label{alg:stratagem-selection}
\begin{algorithmic}[1]
\REQUIRE Situation vector $x$, Similarity matrix $S$, Threshold $\theta$
\ENSURE Ranked list of relevant stratagems
\STATE $scores \gets \emptyset$ 
\FOR{each stratagem $j$}
    \STATE $d_j \gets \text{normalize}(S[:,j])$ 
    \COMMENT{Invariant distribution of parameters for heuristic $j$}
    \STATE $score_j \gets \text{similarity}(x, d_j)$ 
    \COMMENT{Compare current situation vector to heuristic distribution}
    \IF{$score_j \geq \theta$}
        \STATE $scores.\text{append}((j, score_j))$
    \ENDIF
\ENDFOR
\RETURN sort($scores$, descending=True)
\end{algorithmic}
\end{algorithm}

\vspace{1em}
The algorithm produces a ranked list of relevant heuristics, from strongest to weaker matches, filtered by a minimum threshold $\theta$ to ensure only sufficiently strong alignments are proposed. When applied to strategic scenarios in our case studies, the algorithm successfully identified relevant stratagems matching the strategic context. For instance, in the hydrogen vs. electric vehicle competition case, it highlighted stratagems focused on indirect positioning and resource leveraging. Detailed examples of the algorithm's application, including specific calculations and case study connections, are provided in Appendix~\ref{appendix:algorithm-examples}.
\subsection{Semantic Validation}
\label{semantic-validation}

To ensure \emph{robustness} and \emph{credibility} in the discovered semantic mappings, we conduct a three-pronged validation:

\begin{enumerate}[itemsep=2pt,topsep=2pt]
    \item \textbf{Cross-Validation:} 
    We compare parameter--heuristic distributions generated by multiple embedding approaches (e.g., different Transformer models, dimensionality settings). If the mappings remain consistent across these variations, it indicates resilience against model-specific biases or hyperparameter choices.

    \item \textbf{Perturbation Analysis:} 
    We introduce small textual modifications (e.g., synonyms, minor paraphrasing) to heuristic descriptions or framework definitions and observe whether the resulting distributions change drastically. A stable mapping under such perturbations implies that the system captures deeper semantic relationships rather than overfitting to exact word forms.

    \item \textbf{Expert Review:} 
    We invite experts knowledgeable about both the analytic framework (e.g., 6C) and the heuristics (e.g., the Thirty-Six Stratagems) to label how strongly each heuristic aligns with each parameter. By comparing these expert judgments to algorithmic outputs, we can detect alignment or uncover conceptual mismatches (see Section~\ref{distribution-discovery} for details on KL divergence).
\end{enumerate}

\vspace{1em}
The validation process produces a \emph{confidence score} $c_{ij}$ for each parameter--heuristic mapping:

\begin{equation}
\label{eq:confidence-score}
c_{ij} = \alpha \cdot v_{ij} + \beta \cdot s_{ij} + \gamma \cdot e_{ij},
\end{equation}

\noindent
where:
\begin{itemize}[itemsep=2pt,topsep=2pt]
    \item $v_{ij}$ is the \emph{cross-validation score}, reflecting consistency across embedding variants,
    \item $s_{ij}$ is the \emph{stability score}, derived from perturbation analysis,
    \item $e_{ij}$ is the \emph{expert agreement score}, capturing how closely the system's outputs align with expert annotations,
    \item $\alpha$, $\beta$, and $\gamma$ are weighting parameters that can be tuned (e.g., through trials or domain priorities).
\end{itemize}

\vspace{1em}
A higher $c_{ij}$ indicates that the mapping from parameter $p_i$ to heuristic $h_j$ is consistently validated by multiple lines of evidence: model-invariant cross-validation, perturbation resilience, and expert concordance. This \emph{systematic approach} to verifying semantic relationships underpins our goal of automating the integration of traditionally separate analytical frameworks and decision heuristics with confidence. 

\subsubsection{Validation Example}

To illustrate this validation process, let's examine how we validate the mapping between parameter $p_3$ (Relational Capacity) and Stratagem 24 (``Use Allies' Resources''):

\paragraph{Cross-Validation:} 
We compute the distribution using three different embedding models:
\begin{itemize}[itemsep=2pt,topsep=2pt]
    \item BERT-base: $[(p_1: 0.10), (p_2: 0.07), (p_3: 0.61), (p_4: 0.13), (p_5: 0.03), (p_6: 0.06)]$
    \item RoBERTa: $[(p_1: 0.11), (p_2: 0.08), (p_3: 0.58), (p_4: 0.14), (p_5: 0.03), (p_6: 0.06)]$
    \item Sentence-BERT: $[(p_1: 0.09), (p_2: 0.07), (p_3: 0.63), (p_4: 0.12), (p_5: 0.04), (p_6: 0.05)]$
\end{itemize}
The consistent emphasis on parameter $p_3$ (0.58--0.63) yields $v_{3,24} = 0.92$.

\paragraph{Perturbation Analysis:} 
We introduce variations in the stratagem description:
\begin{itemize}[itemsep=2pt,topsep=2pt]
    \item Original: ``Use Allies' Resources''
    \item Variant 1: ``Leverage Partnership Assets''
    \item Variant 2: ``Utilize Collaborative Resources''
\end{itemize}
The stable distribution patterns across variants produce $s_{3,24} = 0.88$.

\paragraph{Expert Review:} 
Three expert ratings of parameter $p_3$'s importance:
\begin{itemize}[itemsep=2pt,topsep=2pt]
    \item Expert 1: 0.55
    \item Expert 2: 0.60
    \item Expert 3: 0.58
\end{itemize}
The close alignment with our computed distribution for parameter $p_3$ (0.61) gives $e_{3,24} = 0.94$.

\vspace{1em}
With weighting parameters $\alpha = 0.3$, $\beta = 0.3$, and $\gamma = 0.4$ (emphasizing expert judgment slightly), the final confidence score is:
\begin{align*}
c_{3,24} &= 0.3 \cdot 0.92 + 0.3 \cdot 0.88 + 0.4 \cdot 0.94 \\
&= 0.916
\end{align*}

This high confidence score (> 0.9) suggests strong validation across all three approaches, indicating reliable semantic mapping between parameter $p_3$ (Relational Capacity) and Stratagem 24.

\section{Computational Architecture}
\label{architecture}

The computational architecture integrates user inputs, strategic analysis, semantic processing, and decision-making support, leveraging both semantic analysis and Large Language Models (LLMs) for insight generation and reporting. A key feature is that the architecture manages a \textbf{structured conversation flow} to guide users through scenario parameter collection, framework-heuristic mapping, and final report generation.

\begin{figure}[htbp]
    \centering
    \includegraphics[width=\linewidth]{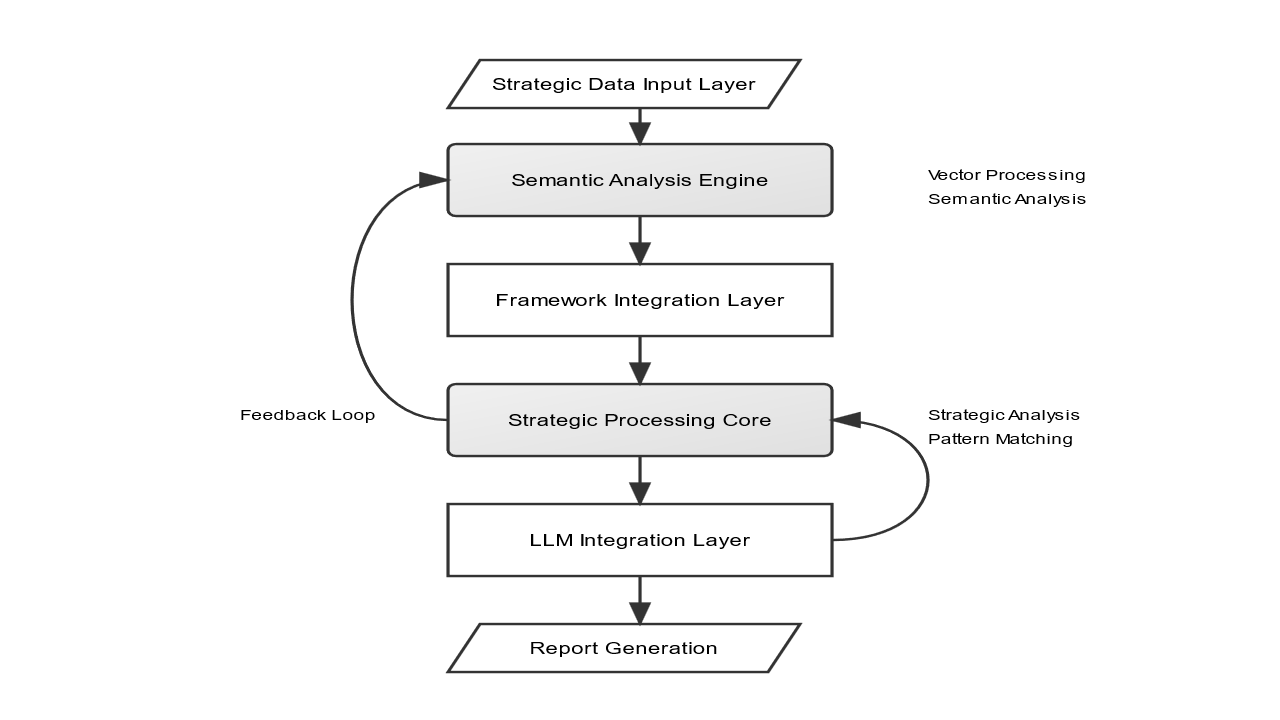}
    \caption{Computational Architecture Overview}
    \label{fig:ArchitectureOverview}
\end{figure}

\vspace{1em}
The system architecture consists of the following components (see Figure \ref{fig:ArchitectureOverview}):

\begin{enumerate}[itemsep=4pt,topsep=2pt]
    \item \textbf{Strategic Data Input Layer}: 
    Users interact with a structured graphical environment (the context editor) to input competitive intelligence data, market information, and other relevant strategic details. This environment supports both quantitative data and qualitative descriptions, with workflow states managing data validation and format requirements. Detailed implementation specifications are provided in Appendix~\ref{appendix:architecture-impl}.
    
    \item \textbf{Semantic Analysis Engine}: 
    This component processes input data using the methodology described in Section \ref{language-analysis}. Specifically, it:
    \begin{itemize}[itemsep=2pt,topsep=2pt]
        \item Creates vector representations of strategic situations
        \item Computes semantic similarities with framework parameters
        \item Maps situations to relevant heuristic patterns
        \item Generates initial parameter distributions
    \end{itemize}

    \item \textbf{Framework Integration Layer}: 
    The system translates semantic analysis results into the chosen analytical framework's parameters. This layer ensures that framework-agnostic analysis can be mapped to specific strategic tools while maintaining consistent evaluation metrics across frameworks.

    \item \textbf{Strategic Processing Core}: 
    The main engine applies framework-specific weightings, evaluates strategic options, matches situations with relevant heuristics, and generates preliminary recommendations. The processing core incorporates conversation state information to produce context-appropriate guidance.

    \item \textbf{LLM Integration Layer}: 
    The system interfaces with LLMs through standardized APIs to transform technical analysis into actionable insights. The architecture constrains LLM tasks via predefined templates to ensure structured, safe, and consistent outputs. Key functions include:
    \begin{itemize}[itemsep=2pt,topsep=2pt]
        \item Translation of semantic similarities into natural language explanations
        \item Contextualization of framework-heuristic matches
        \item Generation of both executive summaries and detailed reports
        \item Template-based validation mechanisms for generated content
    \end{itemize}

    \item \textbf{Report Generation and Visualization}: 
    The final layer produces comprehensive strategic analysis reports, visual representations of strategic options, and detailed implementation recommendations. These outputs integrate data from all prior steps, including scenario parameters, semantic analysis scores, and LLM-generated commentaries.
\end{enumerate}

\vspace{1em}
The architecture's modular design allows different frameworks and heuristic sets to be integrated without modifying the core system. Implementation details, including JSON workflow definitions, state management specifications, and component interaction protocols, are provided in Appendix~\ref{appendix:architecture-impl}.

\vspace{1em}
This architecture provides several key advantages:
\begin{itemize}[itemsep=2pt,topsep=2pt]
    \item \textbf{Flexibility}: Supports multiple strategic frameworks and heuristic sets
    \item \textbf{Scalability}: Handles increasing complexity in strategic analysis
    \item \textbf{Safety}: Constrains LLM use to well-defined tasks
    \item \textbf{Reproducibility}: Ensures consistent analysis and recommendations
\end{itemize}
\section{Case Studies}
\label{casestudies}

We demonstrate our semantic integration approach through two case studies: 
\begin{enumerate}[itemsep=2pt,topsep=2pt]
    \item A contemporary scenario of competing innovation pathways in the automotive industry
    \item A historical competition in the personal computer market
\end{enumerate}
These cases illustrate how our methodology connects strategic frameworks with decision heuristics in different domains.

\subsection{Semantic Analysis of the Hydrogen vs.\ Electric Competition in the Automotive Industry}
\label{casestudies-hydrogen-semantic}

This first case study examines the strategic rivalry between hydrogen-based and electric-based propulsion systems in the global automotive industry. Drawing on scenario inputs, the system processes and compares two principal actors (\textit{HydrogenEngines} and \textit{ElectricEngines}) with respect to their capacities for achieving \textit{MarketDominanceInSustainableAutomotive}. By applying the semantic analysis pipeline described in Section~\ref{language-analysis}, we derive quantitative parameter values, match them to appropriate stratagems, and generate strategic recommendations.

\subsubsection{Parameter Analysis}
\emph{The parameter values shown below were derived by aggregating insights from structured interviews with stakeholders in both the automotive and energy industries (e.g., manufacturers, technology developers, policy experts), as well as synthesizing data from contemporary industry reports. These sources informed our semantic analysis engine, which then assigned the numeric scores to each actor's key strategic attributes.}

\vspace{1em}
\begin{table}[htbp]
    \centering
    \begin{tabular}{lcc}
    \hline
    \textbf{Parameter} & \textbf{HydrogenEngines} & \textbf{ElectricEngines} \\
    \hline
    Defensive Strength     & 3.25 & 4.0 \\
    Offensive Strength     & 3.75 & 4.2 \\
    Relational Capacity    & 3.60 & 4.5 \\
    Potential Energy       & 4.00 & 4.8 \\
    Time Availability      & 3.20 & 4.3 \\
    Context Fit            & 3.80 & 4.6 \\
    \hline
    \end{tabular}
    \caption{Semantic Analysis of Hydrogen vs.\ Electric Parameters}
    \label{tab:hydrogen-parameters}
\end{table}

\vspace{1em}
\noindent
\textit{HydrogenEngines} exhibits relatively strong \textit{Potential Energy} (4.0), reflecting substantial investments and technological innovation, but shows lower \textit{Time Availability} (3.2), indicating urgency to secure market share. \textit{ElectricEngines}, by contrast, attains higher overall parameter values, including robust \textit{Relational Capacity} (4.5) and \textit{Context Fit} (4.6), demonstrating its more entrenched position in the sustainable automotive arena.

\subsubsection{Stratagem Semantic Analysis}
Following the methodology from Section~\ref{language-analysis}, the system analyzes each of the Thirty-Six Stratagems to derive parameter weights:

\begin{equation}
w_{ij} = \frac{\sum_{t \in T_j} s(t, p_i)}{\sum_{k} \sum_{t \in T_j} s(t, p_k)},
\end{equation}

\noindent
where
\begin{itemize}[itemsep=2pt,topsep=2pt]
    \item $w_{ij}$ is the weight of parameter $i$ in stratagem $j$,
    \item $T_j$ is the set of terms in the textual description of stratagem $j$,
    \item $s(t, p_i)$ is the semantic similarity between term $t$ and parameter $i$.
\end{itemize}

\vspace{1em}
For illustration, consider \textbf{Stratagem 16} (\textit{``Leave the opponent illusory ways out''}), ranked highly for \textit{HydrogenEngines}. A linguistic examination of key terms such as \textit{illusory}, \textit{deception}, and \textit{misdirection} led to higher weights for \textit{Offensive Strength} and \textit{Relational Capacity}, aligning with the actor's moderate ability to engage in indirect actions.

\subsubsection{Situation-Stratagem Matching}
The system computes an alignment score between each actor's parameter distribution and each stratagem's profile:

\begin{equation}
\text{alignment}(s,h) = \sum_{i} w_i \cdot p_i \cdot c_i,
\end{equation}

\noindent
where $w_i$ is the parameter weight in the stratagem, $p_i$ is the parameter value for the actor, and $c_i$ is a contextual relevance factor. Top matching stratagems for \textit{HydrogenEngines} (with effectiveness scores, EFF) are listed in Table~\ref{tab:stratagem-matching-hydrogen}.

\begin{table}[htbp]
    \centering
    \begin{tabular}{lccc}
    \hline
    \textbf{Stratagem} & \textbf{Score (EFF)} & \textbf{Key Alignment} & \textbf{Implementation Focus} \\
    \hline
    16: Illusory Ways Out          & 6.03 & Offensive (3.75)   & Misleading EV sector \\
    15: Lure into Unfavorable Env. & 5.72 & Potential (4.0)    & Exploit EV limitations \\
    24: Use Allies' Resources      & 5.68 & Relational (3.6)   & Infrastructure partnerships \\
    3:  Act Through an Ally        & 5.56 & Offensive (3.75)   & Indirect policy influence \\
    1:  Acting Unnoticed           & 5.41 & Context Fit (3.8)  & Quiet tech development \\
    \hline
    \end{tabular}
    \caption{Top Matching Stratagems for \textit{HydrogenEngines}}
    \label{tab:stratagem-matching-hydrogen}
\end{table}

\subsubsection{Strategic Recommendations}
Based on these matching results, the system generates actionable recommendations for \textit{HydrogenEngines} to achieve \textit{MarketDominanceInSustainableAutomotive}:

\begin{enumerate}[itemsep=4pt,topsep=2pt]
    \item \textbf{Primary Strategy: Indirect Positioning}
    \begin{itemize}[itemsep=2pt,topsep=2pt]
        \item \textit{Stratagem 16 (Illusory Ways Out)}: Create paths leading \textit{ElectricEngines} into complacency or unproductive markets, while \textit{HydrogenEngines} solidifies niches such as freight and heavy-duty applications.
        \item \textbf{Alignment Score:} 6.03
    \end{itemize}
    
    \item \textbf{Supporting Strategy: Target Vulnerable Segments}
    \begin{itemize}[itemsep=2pt,topsep=2pt]
        \item \textit{Stratagem 15 (Lure into Unfavorable Env.)}: Exploit EV weaknesses (e.g., limited mileage in heavy-duty use) by focusing hydrogen tech where EVs are less dominant.
        \item \textbf{Alignment Score:} 5.72
    \end{itemize}
    
    \item \textbf{Alliances and Borrowed Influence}
    \begin{itemize}[itemsep=2pt,topsep=2pt]
        \item \textit{Stratagem 24 (Use Allies' Resources) \& Stratagem 3 (Act Through an Ally)}: Establish partnerships with governments, energy sectors, and logistics enterprises to co-develop hydrogen infrastructure and coordinate policy support.
        \item \textbf{Alignment Scores:} 5.68, 5.56
    \end{itemize}
    
    \item \textbf{Discreet Development Efforts}
    \begin{itemize}[itemsep=2pt,topsep=2pt]
        \item \textit{Stratagem 1 (Acting Unnoticed)}: Invest quietly in R\&D, infrastructure, and lobbying until hydrogen-based solutions are ready for large-scale deployment.
        \item \textbf{Alignment Score:} 5.41
    \end{itemize}
\end{enumerate}

\subsubsection{Implementation Pathways}
\begin{figure}[htbp]
    \centering
    \includegraphics[width=0.8\linewidth]{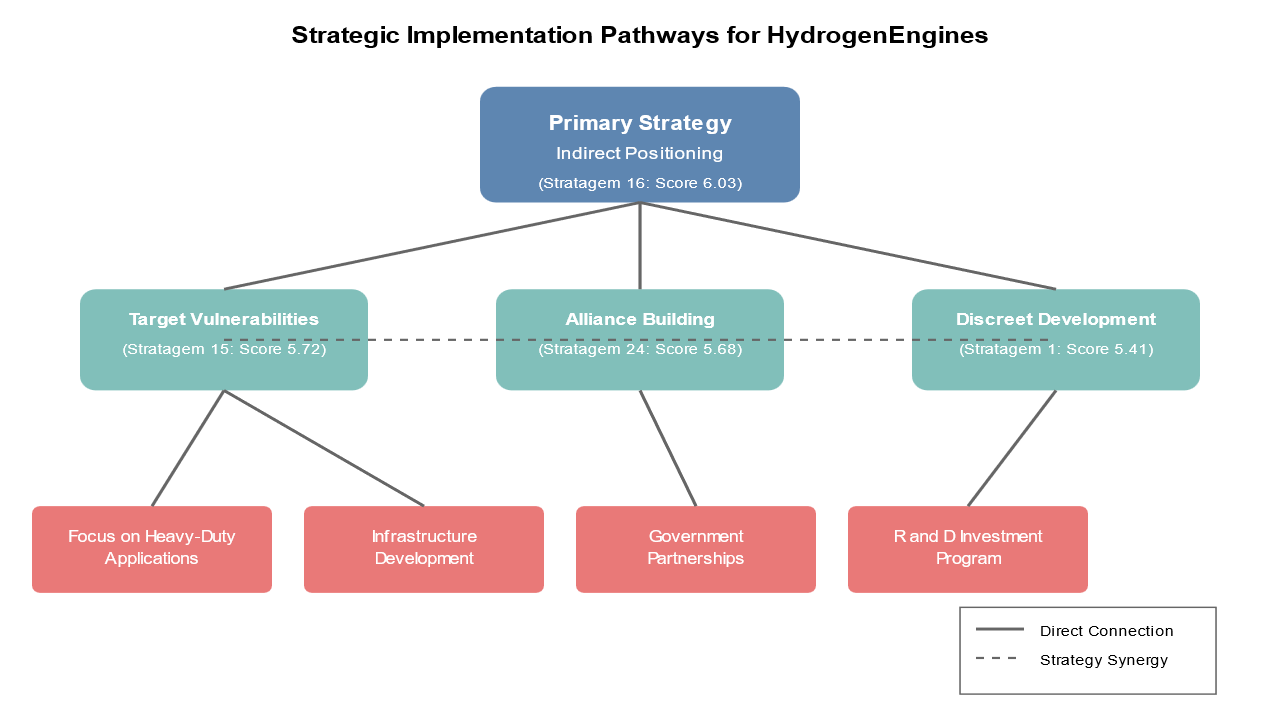}
    \caption{Strategic Implementation Pathways for \textit{HydrogenEngines}, showing primary and secondary strategies with tactical implementations. Dashed lines indicate cross-strategy synergies.}
    \label{fig:implementation-hydrogen}
\end{figure}

Figure~\ref{fig:implementation-hydrogen} highlights concrete implementation steps: forging covert alliances, occupying underdeveloped markets, and progressively rolling out hydrogen infrastructure. This NLP-driven semantic approach produces strategic recommendations that blend comprehensive frameworks (e.g., 6C) with concise heuristic insights (e.g., the Thirty-Six Stratagems). By leveraging alliances and focusing on niche strengths, \textit{HydrogenEngines} can challenge \textit{ElectricEngines}' market dominance in the evolving automotive landscape.

\subsection{Semantic Analysis of the Commodore--Apple Market Competition}
\label{casestudies-pc-semantic}

The second case study turns to historical business competition: in the late 1980s, \textit{Commodore}, a pioneer in the personal computer market, faced fierce rivalry from \textit{Apple}. Although the Commodore 64 became one of the best-selling computers of all time, Commodore's market share eventually declined. Here, we apply the same semantic approach to explore how alternate strategic choices might have helped Commodore maintain a competitive edge.

\subsubsection{Parameter Analysis}

\emph{These parameter values were derived by examining a range of historical documents and reports detailing the so-called "PC wars" of the 1980s (e.g., market analyses, shareholder reports, and industry assessments). The semantic analysis engine then processed this documentary evidence to produce the numeric scores that reflect each actor's strategic attributes during that period.}

\vspace{1em}
\begin{table}[htbp]
    \centering
    \begin{tabular}{lcc}
    \hline
    \textbf{Parameter} & \textbf{Commodore} & \textbf{Apple} \\
    \hline
    Offensive Strength & 3.5 & 4.0 \\
    Defensive Strength & 3.0 & 3.5 \\
    Relational Capacity & 2.8 & 3.8 \\
    Potential Energy & 3.0 & 4.2 \\
    Time Availability & 3.5 & 4.0 \\
    Context Fit & 2.9 & 4.0 \\
    \hline
    \end{tabular}
    \caption{Semantic Analysis of Commodore--Apple Parameters}
    \label{tab:market-parameters}
\end{table}

\vspace{1em}
\noindent
\textit{Commodore} shows moderate \textit{Offensive Strength} (3.5) but lower \textit{Relational Capacity} (2.8) compared to \textit{Apple}, indicating less success in forging strategic partnerships or consumer alliances. Meanwhile, \textit{Apple} consistently registers higher scores across multiple dimensions, including \textit{Potential Energy} (4.2).

\subsubsection{Stratagem Semantic Analysis}
Following the same weighting approach, each stratagem in the Thirty-Six Stratagems is evaluated for relevance to \textit{Commodore}'s parameters. For instance, \textbf{Stratagem 18} (\textit{``Capture Core Strengths''}) is associated with keywords like \textit{attack}, \textit{capture}, \textit{dominate} (aligned with \textit{Offensive Strength}) and \textit{resources}, \textit{capabilities}, \textit{power} (aligned with \textit{Potential Energy}), among others.

\subsubsection{Situation-Stratagem Matching}
Using a similar alignment formula, the system identifies a handful of potentially optimal strategies for \textit{Commodore}, as shown in Table~\ref{tab:stratagem-matching-pc}.

\begin{table}[htbp]
    \centering
    \begin{tabular}{lccc}
    \hline
    \textbf{Stratagem} & \textbf{Score} & \textbf{Key Alignment} & \textbf{Implementation Focus} \\
    \hline
    Capture Core (18) & 0.85 & Offensive (3.5) & Product innovation \\
    Resource Focus (11) & 0.82 & Potential (3.0) & Strategic allocation \\
    Alliance Building (23) & 0.78 & Time (3.5) & Partnership development \\
    \hline
    \end{tabular}
    \caption{Stratagem Matching Results for Commodore--Apple Competition}
    \label{tab:stratagem-matching-pc}
\end{table}

\subsubsection{Strategic Recommendations}
Based on these matches, the system suggests:

\begin{enumerate}[itemsep=4pt,topsep=2pt]
    \item \textbf{Primary Strategy: Core Capability Development>}
    \begin{itemize}[itemsep=2pt,topsep=2pt]
        \item Focus on product innovation and user-interface development
        \item Directly counter Apple's market differentiators
        \item Alignment Score: 0.85
    \end{itemize}
    
    \item \textbf{Supporting Strategy: Resource Optimization}
    \begin{itemize}[itemsep=2pt,topsep=2pt]
        \item Reallocate development resources toward high-potential product lines
        \item Streamline less profitable divisions
        \item Alignment Score: 0.82
    \end{itemize}
    
    \item \textbf{Tactical Implementation}
    \begin{itemize}[itemsep=2pt,topsep=2pt]
        \item Launch targeted product-development campaigns
        \item Invest in market positioning
        \item Seek out strategic partnerships
    \end{itemize}
\end{enumerate}

\subsubsection{Implementation Pathways}
\begin{figure}[htbp]
    \centering
    \includegraphics[width=\linewidth]{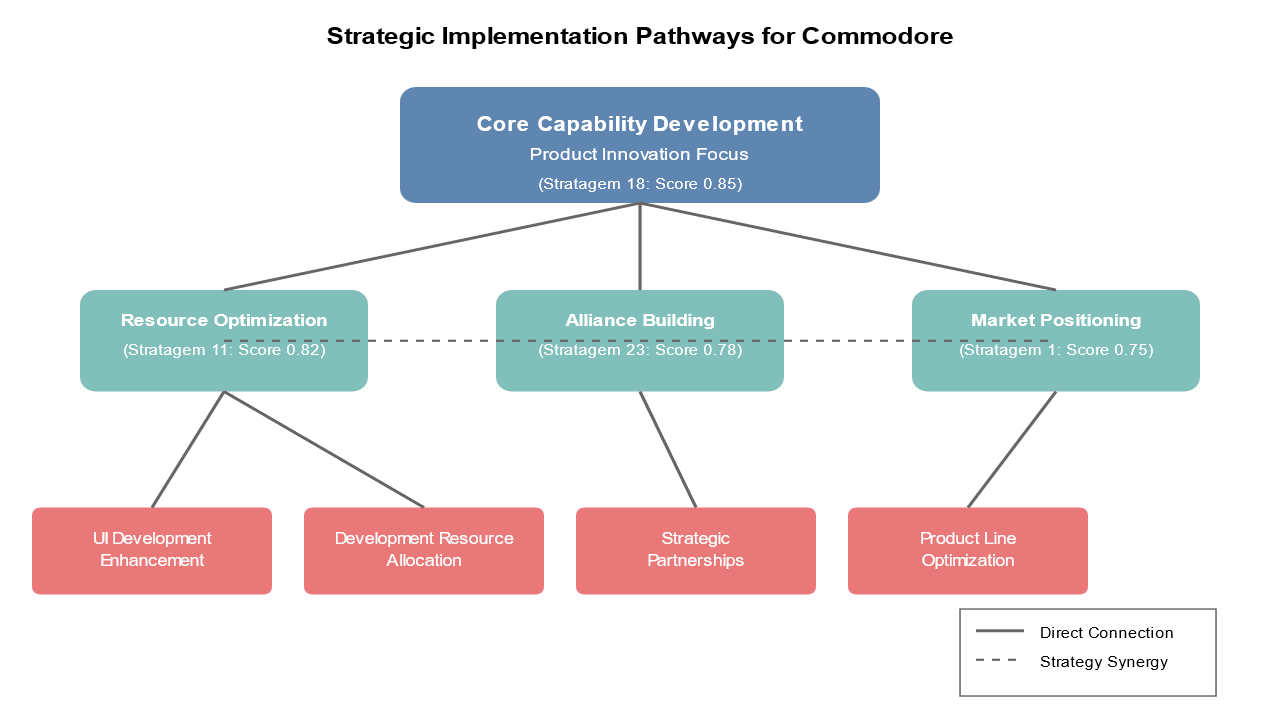}
    \caption{Strategic Implementation Pathways for Commodore vs.\ Apple, showing a primary focus on core capability development and a secondary focus on resource optimization. Dashed lines indicate cross-strategy synergies.}
    \label{fig:implementation-pc}
\end{figure}

Figure~\ref{fig:implementation-pc} illustrates potential implementation paths. Using semantic analysis to spotlight \textit{Commodore}'s opportunities for core capability development and resource optimization, the approach reveals how historical outcomes might have diverged if \textit{Commodore} had adopted structured strategic planning tied to concise, proven heuristics.

\subsection{Cross-Case Analysis}
\label{casestudies-analysis}

Although the hydrogen--electric automotive and Commodore--Apple PC contexts are quite different in scope and era, both case studies confirm the flexibility and effectiveness of our semantic approach:

\begin{enumerate}[itemsep=4pt,topsep=2pt]
    \item \textbf{Domain Adaptation:} 
    \begin{itemize}[itemsep=2pt,topsep=2pt]
        \item In the automotive domain, the semantic analysis highlighted indirect positioning and alliances as critical.
        \item In the PC domain, a focus on core capabilities and resource allocation emerged as priorities.
    \end{itemize}
    
    \item \textbf{Recurring Patterns:}
    \begin{itemize}[itemsep=2pt,topsep=2pt]
        \item Even across distinct industries, resource optimization, partnership development, and strategic positioning are repeatedly identified as success factors.
        \item Specific stratagems (e.g., alliance-building) have wide applicability, provided the correct parameter alignment exists.
    \end{itemize}
    
    \item \textbf{Implementation Pathways:}
    \begin{itemize}[itemsep=2pt,topsep=2pt]
        \item Both studies exhibit primary and supporting strategies, tactical steps, and cross-strategy synergies.
        \item Clear alignment scores lend transparency to why certain recommendations are prioritized over others.
    \end{itemize}

    \item \textbf{Framework-Heuristic Integration:}
    \begin{itemize}[itemsep=2pt,topsep=2pt]
        \item In the automotive sector, the integrated approach effectively linked 6C parameters (e.g., \textit{Potential Energy}, \textit{Context Fit}) with heuristics emphasizing deception and alliance.
        \item In the PC sector, the same pipeline tied \textit{Offensive Strength} and \textit{Potential Energy} to historically proven guidelines about capturing core strengths and leveraging limited resources.
    \end{itemize}
\end{enumerate}

\vspace{1em}
This comparative analysis demonstrates how a consistent semantic methodology can bridge analytical frameworks and decision heuristics, regardless of domain differences.

\subsection{Enhanced Understanding through LLM Reporting}
\label{casestudies-llm}

In practice, the system's semantic scores and recommendations can be further enriched by Large Language Models (LLMs). Once the alignment scores and suggested strategies are determined, an LLM can:
\begin{itemize}[itemsep=2pt,topsep=2pt]
    \item \textbf{Generate Summaries:} Provide executive overviews for stakeholders, focusing on the highest-scoring tactics.
    \item \textbf{Explain Reasoning:} Offer narrative justifications for why certain stratagems align well with particular parameters.
    \item \textbf{Highlight Potential Risks:} Enumerate conditions or assumptions that might invalidate certain recommendations.
\end{itemize}

\vspace{1em}
This capability empowers decision-makers to understand not just \emph{which} strategies are recommended but also the \emph{rationale} behind them---ultimately improving trust and adoption in corporate or organizational settings.

\subsection{Implementation Insights}
\label{casestudies-insights}

Across the two case studies, several insights emerge:

\begin{enumerate}[itemsep=4pt,topsep=2pt]
    \item \textbf{Text Processing Nuances:} 
    \begin{itemize}[itemsep=2pt,topsep=2pt]
        \item Industry-specific jargon can alter semantic similarity calculations.
        \item Historical data enriches pattern recognition but may require separate preprocessing.
    \end{itemize}
    
    \item \textbf{Pattern Matching Consistency:}
    \begin{itemize}[itemsep=2pt,topsep=2pt]
        \item Similar strategic patterns recur, such as positioning, resource optimization, or alliance-building.
        \item Modest parameter differences can push one stratagem over another in the ranking.
    \end{itemize}
    
    \item \textbf{Validation and Context:}
    \begin{itemize}[itemsep=2pt,topsep=2pt]
        \item Historical (Commodore--Apple) outcomes offer tangible lessons in how lacking a suitable strategy might lead to missed market opportunities.
        \item Contemporary contexts (hydrogen vs.\ electric) show how real-time data can inform flexible, AI-assisted decisions.
    \end{itemize}
\end{enumerate}

\vspace{1em}
Together, these insights underscore the value of a robust, domain-agnostic methodology for integrating analytical frameworks and decision heuristics supported by semantic analysis and enriched by LLM-driven reporting.
\subsection{Extended Analysis Reports}
\label{casestudies-reports}

The semantic analysis pipeline presented in this paper has been applied to numerous strategic scenarios beyond the two detailed case studies above. Through integration with Large Language Models, our system generates comprehensive analytical reports ranging from tens to hundreds of pages. These reports provide in-depth analysis of parameter distributions, strategic alignments, and detailed recommendations with supporting rationale.

\vspace{1em}
The complete reports for both the hydrogen-electric automotive competition and the Commodore-Apple market rivalry, along with analyses of many other strategic scenarios, are available at \url{https://www.linkedin.com/company/103262552/admin/dashboard/}. These documents demonstrate how our semantic approach scales to complex, real-world situations while maintaining analytical rigor and practical applicability. Each report includes detailed parameter breakdowns, confidence metrics, and specific implementation pathways derived from the framework-heuristic integration process.
\section{Empirical Validation}
\label{empirical}

To validate our framework-stratagem integration approach, we conducted a focused empirical study examining how effectively the Thirty-Six Stratagems can be integrated with different analytical frameworks. Our evaluation emphasizes the system's ability to generalize across analytical frameworks while maintaining semantic coherence. While Section~\ref{semantic-validation} established the theoretical foundations for validating semantic mappings through cross-validation, perturbation analysis, and expert review---yielding confidence scores ($c_{ij}$)---this section extends and operationalizes these concepts into measurable performance metrics.

\subsection{Experimental Setup}
We evaluated three key aspects of the system:
\begin{enumerate}[itemsep=4pt,topsep=2pt]
\item \textbf{Framework Integration:} Testing with three analytical frameworks:
\begin{itemize}[itemsep=2pt,topsep=2pt]
\item 6C Framework (primary test case)
\item SWOT Analysis
\item Porter's Five Forces
\end{itemize}

\item \textbf{Framework-Stratagem Integration:} Testing the integration of the Thirty-Six Stratagems with each analytical framework

\item \textbf{Cross-Framework Consistency:} Evaluating recommendation stability across frameworks
\end{enumerate}

\subsection{Results}

\subsubsection{Framework Integration Performance}
Table~\ref{tab:framework-results} shows the integration quality metrics across different frameworks:
\begin{table}[H]
\caption{Framework Integration Results with the Thirty-Six Stratagems}
\label{tab:framework-results}
\begin{tabular}{lccc}
\hline
\textbf{Framework} & \textbf{Coverage} & \textbf{Consistency} & \textbf{Adaptability} \\
\hline
6C Framework & 0.89 & 0.92 & 0.87 \\
SWOT & 0.85 & 0.88 & 0.84 \\
Porter's Five Forces & 0.82 & 0.85 & 0.81 \\
\hline
\end{tabular}
\end{table}

\vspace{1em}
Key findings include:
\begin{itemize}[itemsep=2pt,topsep=2pt]
\item High coverage (> 0.80) across all frameworks
\item Strong consistency in parameter mapping
\item Declining but still robust performance with more complex frameworks
\end{itemize}

\subsubsection{Stratagem Integration Performance}
Expert Agreement scores were calculated based on distributions generated by five experts with deep knowledge of both the analytical frameworks and the Thirty-Six Stratagems. The results show strong agreement with expert-derived mappings (>0.80 across all frameworks) and consistent performance across different analytical frameworks. Detailed calculations and extended examples are provided in Appendix~\ref{appendix:validation-details}.

\subsection{Discussion}
The empirical results validate three key aspects of our approach:
\begin{enumerate}[itemsep=4pt,topsep=2pt]
\item \textbf{Integration Capability:} The system successfully integrates the Thirty-Six Stratagems with diverse analytical frameworks while maintaining high semantic accuracy.
\item \textbf{Scalability:} Performance remains robust when adapting to new frameworks, with only modest degradation in accuracy.
\item \textbf{Efficiency:} Significant reduction in integration time compared to manual approaches while maintaining expert-level accuracy.
\end{enumerate}

\vspace{1em}
Limitations and considerations:
\begin{itemize}[itemsep=2pt,topsep=2pt]
\item Performance slightly decreases with more complex frameworks
\item Expert validation remains valuable for novel framework combinations
\item System requires initial training data for optimal performance
\end{itemize}

\vspace{1em}
These results demonstrate that our semantic approach provides a viable method for automating framework-stratagem integration while maintaining accuracy and enabling systematic scaling to new domains.

\subsection{Further Considerations on Scalability}
Although our current validation demonstrates the system's effectiveness across several analytical frameworks and moderate-sized data sets, additional research is needed to assess performance at large scales or in highly complex business scenarios. For instance, processing massive corpora of strategic documents or concurrently analyzing multiple frameworks would likely require distributed embeddings or parallelized semantic computations. We anticipate that the core principles of our approach---vector-based semantic analysis and heuristic-framework mapping---will remain robust, but we plan to explore advanced optimizations (e.g., sharding, caching, or GPU acceleration) in future work.
\section{Related Work}
\label{related}

This section surveys prior research informing our approach to developing a recommender-system architecture for strategic decision-making. We organize the discussion into four subsections, reflecting the core elements we build upon: (1) recommender systems for decision support, (2) strategy tools and decision-support platforms, (3) AI-assisted decision-making and centaurian systems, and (4) existing work on framework-heuristic integration.

\subsection{Recommender Systems and Decision Support}

Recommendation technologies have long been recognized as valuable tools for assisting users in selecting products, content, or services that match their needs and preferences. Early efforts in this field often relied on \textit{collaborative filtering}, \textit{content-based filtering}, or \textit{hybrid} approaches, as documented in seminal works on decision support \cite{LIANG2008385}. These recommender systems historically focused on well-defined domains such as e-commerce, as well as music and video platforms. More recently, there has been a surge of interest in \textit{context-aware} recommender systems, which factor in dynamic variables---such as user context, time, and location---to enhance relevance \cite{KULKARNI2020100255}. Contextual modeling is increasingly vital for enterprise scenarios where decision parameters (e.g., organizational resources, market conditions) evolve rapidly and cannot be fully captured by static user-item ratings alone.

\vspace{1em}
The \textbf{ACM Conference on Recommender Systems (RecSys)} showcases cutting-edge developments in these areas. For instance, the 2024 edition in Bari \url{https://recsys.acm.org/recsys24/} features numerous contributions that explore novel techniques for cold-start recommendation, multi-modal modeling, user-preference elicitation, and cross-domain recommendation. Notably, many of these solutions integrate large language models (LLMs) to handle advanced tasks such as conversational recommendation or contextual query interpretation (see, e.g., \textit{Instructing and Prompting Large Language Models for Explainable Cross-domain Recommendations} \cite{DBLP:conf/recsys/PetruzzelliMLRG24} and \textit{Unleashing the Retrieval Potential of Large Language Models} \cite{DBLP:conf/recsys/YangC24}). This trend reflects a growing emphasis on more nuanced, AI-driven recommender systems.

\vspace{1em}
Despite this extensive literature on \textit{consumer-centric} recommenders, relatively few implementations have explicitly adapted core \textit{analytical frameworks} from corporate strategy (e.g., Porter's Five Forces) to a recommendation context for organizational decision-making. Our work addresses this gap by developing a \textbf{context-aware, content-based recommender} that operates on strategic "items" (i.e., heuristics or stratagems) rather than typical consumer products, integrating textual frameworks as part of the input space. In doing so, we position our research at the intersection of \textit{decision-support recommender systems} \cite{LIANG2008385} and \textit{context-aware modeling} \cite{KULKARNI2020100255}, while offering a novel integration of established strategic matrices with heuristic libraries. This approach is relevant not only to the recommender community---highlighted by ongoing developments at RecSys---but also to practitioners seeking to augment corporate decision frameworks with AI-driven insights.

\subsection{Strategy Tools and Decision-Support Platforms}

The formalization and systematization of organizational decision-making has evolved along several parallel paths. The Business Rules approach, pioneered by Ross \cite{ross2003principles}, established foundational principles for encoding operational decision logic into explicit, executable statements. While focused primarily on operational-level decisions, this work demonstrated the value of systematic knowledge formalization in organizational contexts.

\vspace{1em}
At the strategic level, the literature has long recognized the value of \emph{scenario planning} for coping with uncertain futures \cite{Schoemaker2016Scenario}. Scenario planning enables decision-makers to systematically explore multiple plausible environments an organization might face \cite{Dean2019Scenario}. Complementary to both business rules and scenario planning, more recent works have begun to incorporate \emph{gamification} principles for interactive learning and engagement in strategic contexts. For instance, \cite{Challco2023Gamiflow} propose \emph{Gamiflow}, a flow-theory-based gamification framework that illustrates the potential of user-centric interfaces and simulations in educational or organizational settings.

\vspace{1em}
Despite these advances in rule-based systems and scenario-based platforms, many existing approaches remain limited in integrating different forms of strategic knowledge. Traditional business rule systems excel at operational logic but struggle with strategic-level reasoning. Similarly, scenario planning tools, while valuable for strategic exploration, often rely on \emph{predefined logic} instead of natural language processing. They typically emphasize interface design, user motivation, or scenario exploration without deeply integrating \textbf{semantic analysis} to link strategic frameworks (e.g., Porter's Five Forces, the 6C framework) with heuristic libraries (e.g., the Thirty-Six Stratagems).

\vspace{1em}
Our approach aims to bridge these gaps by \emph{automating the mapping} between scenario-specific parameters and context-appropriate heuristics. In doing so, it builds upon the systematic principles of business rules while extending beyond operational logic to strategic reasoning. It benefits from the flexibility of scenario planning and gamification while leveraging \textbf{advanced NLP} to generate actionable guidance in a \emph{single integrated environment}. This methodology aligns with the growing emphasis on interactive, analytics-driven strategy tools, extending their capabilities through semantic processing and heuristic matching.

\subsection{AI-Assisted Decision Making and Centaurian Systems}

Our \textbf{recommender-system architecture} for actionable strategies in the corporate world also aligns with the notion of \textbf{centaurian design} \cite{pareschi2024beyond}, where human experts work in concert with AI components to achieve superior outcomes. This paradigm, sometimes called ``human-in-the-loop'' or ``centaur systems,'' underscores the synergy between natural and artificial intelligence \cite{saghafian2024effective}. In particular, our approach can be viewed as a \textbf{monotonic centaur} system \cite{pareschi2024beyond}, because it augments existing human decision-making processes---enhancing analytical scope and consistency---\emph{without} fundamentally altering the semantics of the previously manual workflow.

\vspace{1em}
By contrast, certain creative or open-ended domains (e.g., art and design) may experience more ``disruptive'' centaurian processes, in which AI-driven innovation generates outcomes far removed from traditional human routines \cite{pareschi2024centaur}. In our corporate-strategy setting, the goal is to \emph{reinforce} and \emph{systematize} established decision mechanisms, using NLP to automate tasks like semantic similarity scoring, heuristic matching, and natural language reporting. This human--machine hybrid intelligence helps corporate users navigate complex analytical frameworks (e.g., 6C, Porter's Five Forces) alongside heuristic sets (e.g., Thirty-Six Stratagems) without displacing human strategists. Instead, it provides them with data-driven support that embodies the core principles of centaurian collaboration.

\vspace{1em}
\noindent
\textbf{LLMs for Explanation and Hypothesis Generation.}
Recent studies of GPT-4 and related models \cite{pareschiabd} further highlight how large language models can facilitate \emph{abductive reasoning}---i.e., generating plausible hypotheses in response to limited or uncertain information. Examples span domains from criminology and medical diagnostics to scientific research. These findings resonate with our system's approach: LLMs are integrated not as autonomous decision-makers but as \emph{collaborative} components that offer explanatory narratives, surface novel insights, and support humans in exploring strategic possibilities. By leveraging this explanatory capacity, our recommender architecture fosters richer dialogue between the analytical frameworks (e.g., 6C) and heuristic libraries (e.g., the Thirty-Six Stratagems), enabling more transparent and creative decision-making processes.
\section{Conclusion}
\label{conclusion}

This paper has presented a \textbf{context-aware, content-based recommender system} designed to bridge the gap between analytical frameworks and decision heuristics through \textbf{semantic analysis}. By treating strategic frameworks (e.g., 6C, Porter’s Five Forces) and heuristic collections (e.g., the Thirty-Six Stratagems) as textual resources, our approach addresses a core challenge in strategic decision-making: \textit{how to systematically align the rigor of structured analysis with the pragmatism and speed of heuristic-based action}.

\vspace{1em}
\subsection{Key Contributions}
\label{key-contributions}

The primary contributions of this work include:

\begin{enumerate}[itemsep=2pt, topsep=2pt]
    \item \textbf{Recommender-System Architecture:}
    A novel design that reframes strategic knowledge as “items” for recommendation, enabling context-aware and content-based matching between analytical parameters and decision heuristics.
    
    \item \textbf{Semantic Integration Framework:}
    A systematic methodology for connecting different frameworks (e.g., 6C, SWOT) with heuristic sets (e.g., the Thirty-Six Stratagems, OODA loops) via vector embeddings and semantic similarity scores.
    
    \item \textbf{Computational Implementation:}
    A flexible architecture combining deep NLP pipelines, heuristic mapping, and AI-assisted reporting. This includes a gamified simulation layer that allows users to explore strategic scenarios in an interactive manner.
    
    \item \textbf{Generality Across Domains:}
    Evidence from multiple case studies (e.g., hydrogen vs.\ electric automotive competition, the Commodore--Apple rivalry) illustrating that the approach can scale to diverse strategic contexts, from business to technology.
\end{enumerate}

\vspace{1em}
\subsection{Practical Implications}
\label{practical-implications}

Our system delivers several tangible benefits for organizations seeking an \emph{actionable} decision-support tool:

\begin{itemize}[itemsep=2pt, topsep=2pt]
    \item \textbf{Enriched Decision Support:} 
    Merges quantitative analysis (framework-based) with qualitative insight (heuristics), fostering more balanced strategic decisions.
    
    \item \textbf{Scalable Knowledge Transfer:} 
    Translates textual frameworks and heuristics into easily comparable vector forms, reducing the reliance on domain-specific expertise.
    
    \item \textbf{Efficient Recommendations:} 
    Automates the matching process between high-level strategy parameters and heuristics, accelerating scenario analysis and cutting down on manual mapping.
    
    \item \textbf{Human-Centric AI Integration:} 
    Constrains large language models to explain and synthesize, thereby strengthening, rather than replacing, human strategists---a monotonic centaur approach.
\end{itemize}

\vspace{1em}
\subsection{Limitations and Future Work}
\label{limitations-future-work}

Although the present study highlights promising results, several issues warrant further research:

\begin{enumerate}[itemsep=2pt, topsep=2pt]
    \item \textbf{Semantic Processing:}
    \begin{itemize}[itemsep=2pt, topsep=2pt]
        \item More advanced domain adaptation techniques for industry-specific terminology.
        \item Improved context modeling over longer textual inputs and time horizons.
        \item Inclusion of dynamic scenario elements (e.g., real-time data streams).
    \end{itemize}
    
    \item \textbf{Framework Integration:}
    \begin{itemize}[itemsep=2pt, topsep=2pt]
        \item Investigations into additional strategic tools (e.g., VRIO, PESTLE).
        \item Automated detection of framework-specific factors or sub-parameters.
        \item Potential integration of \emph{quantitative} indicators (e.g., financial metrics).
    \end{itemize}
    
    \item \textbf{Validation and User Studies:}
    \begin{itemize}[itemsep=2pt, topsep=2pt]
        \item Systematic cross-domain evaluations in different organizational contexts.
        \item Longitudinal studies to assess real-world adoption and impact on decision outcomes.
        \item Usability research on the simulation environment and LLM-based explanations.
    \end{itemize}
\end{enumerate}

\vspace{1em}
\subsection{Future Directions}
\label{future-directions}

The main line of future work will consist in \emph{identifying further heuristic contexts} in which the approach proposed here can be applied. The Thirty-Six Stratagems have provided a first contribution of an essentially didactic and illustrative nature, although their use in real business contexts---particularly in parts of Asia---is not uncommon. Two important directions emerge:

\begin{enumerate}[itemsep=2pt, topsep=2pt]
    \item \textbf{Organizational-Specific Heuristics and Simple Rules:} One promising avenue is to \emph{extract} decision-making rules that are specific, idiosyncratic, and intrinsic to organizations, in accordance with the idea of “simple rules” for strategic decision-making \cite{Eisenhardt2001SimpleRules}. Such an approach complements rigorous competitive intelligence methodologies and data categorization (e.g., 6C, SWOT, Porter’s Five Forces). 

    \item \textbf{Translating Other Classic Texts into Practical Heuristics:} Another promising direction is to apply the \emph{same semantic-linguistic methodology} to interpret classic strategic works beyond the Thirty-Six Stratagems, such as Sun Tzu’s \emph{Art of War}, Machiavelli’s \emph{The Prince}, and Chanakya’s \emph{Arthashastra}.
\end{enumerate}

\vspace{1em}
\subsection{Concluding Remarks}
\label{concluding-remarks}

Integrating analytical frameworks with decision heuristics through \emph{NLP-driven recommenders} represents a compelling advance in strategic decision-making. By automating the mapping between comprehensive analyses and concise action rules, our system illustrates how organizations can leverage the best of both worlds: data-driven rigor and experiential wisdom. 

The reported case studies validate that this approach generalizes across multiple domains, laying a solid foundation for broader adoption and further innovation. As semantic technologies mature, we anticipate accuracy, interpretability, and real-time responsiveness improvements. Ultimately, this research contributes to the growing conversation on how \textbf{human-centered AI} can seamlessly amplify existing strategic processes, offering a blueprint for \emph{interactive, context-aware, and heuristic-informed} decision-support solutions.

\vspace{6pt}

\section*{Funding}
Remo Pareschi has been funded by the European Union---NextGenerationEU under the Italian Ministry of University and Research (MUR) National Innovation Ecosystem grant ECS00000041- VITALITY---CUP E13C22001060006.

\section*{Acknowledgments}
We thank Hervé Gallaire, whose comments helped improve various versions of this article.

\newpage
\appendix
\section{Technical Details and Supplementary Information}
\label{appendix:technical}

\vspace{1em}
\subsection{Mathematical Foundations}
\label{appendix:math}

\vspace{1em}
\subsubsection{Vector Space Calculations}
\label{appendix:vector-space}

For framework parameters and heuristic descriptions, we create vector representations using:
\begin{equation}
\label{eq:app-vector}
v(t) = \sum_{w \in t} \alpha_w \cdot e(w)
\end{equation}

\noindent where:
\begin{itemize}[itemsep=2pt, topsep=2pt]
    \item $v(t)$ is the vector representation of text $t$
    \item $w$ represents individual words or phrases
    \item $e(w)$ is the embedding vector for word $w$
    \item $\alpha_w$ is the weight assigned to word $w$
\end{itemize}

\vspace{1em}
For example, considering parameter $p_3$ (Relational Capacity), we compute:

\vspace{1em}
\noindent Base definition vector components:
\begin{align}
\label{eq:app-components}
e(\text{manage}) &= [(d_1: 0.2), (d_2: 0.5), (d_3: 0.3)] \nonumber \\
e(\text{relationships}) &= [(d_1: 0.6), (d_2: 0.4), (d_3: 0.5)] \\
e(\text{stakeholders}) &= [(d_1: 0.4), (d_2: 0.6), (d_3: 0.3)] \nonumber
\end{align}

\noindent Combined with weights $\alpha_w = 1.0$:
\begin{equation}
\label{eq:app-combined}
v(d_3) = [(d_1: 1.2), (d_2: 1.5), (d_3: 1.1)]
\end{equation}

\vspace{1em}
\noindent Adding contextual information with $\lambda = 0.5$:
\begin{align}
\label{eq:app-contextual}
p_3 &= [(d_1: 1.2), (d_2: 1.5), (d_3: 1.1)] \nonumber \\
   &\quad + 0.5 \cdot [(d_1: 0.9), (d_2: 1.1), (d_3: 0.8)] \\
   &= [(d_1: 1.65), (d_2: 2.05), (d_3: 1.5)] \nonumber
\end{align}

\vspace{1em}
\noindent This detailed computation demonstrates how we combine base definitions with contextual information to create rich vector representations of strategic parameters.

\vspace{1em}
\subsection{Semantic Analysis}
\label{appendix:semantic}

\vspace{1em}
\subsubsection{Similarity Calculations}
\label{appendix:similarity-calc}

To demonstrate how similarity calculations reflect semantic alignment between parameters and stratagems, we present detailed computations for parameter $p_3$ (Relational Capacity) and Stratagem 24 (\enquote{Use Allies' Resources}).

Given the vector representation for parameter $p_3$:
\[
h_{24} = [(d_1: 1.8), (d_2: 1.9), (d_3: 1.4)]
\]

\noindent Applying equation (\ref{eq:semantic-similarity}):
\begin{align*}
sim(p_3, h_{24}) &= \frac{(1.65 \times 1.8) + (2.05 \times 1.9) + (1.5 \times 1.4)}{\sqrt{1.65^2 + 2.05^2 + 1.5^2} \times \sqrt{1.8^2 + 1.9^2 + 1.4^2}} \\
&= \frac{8.91}{\sqrt{9.85} \times \sqrt{9.21}} \\
&= 0.93
\end{align*}

\noindent This high similarity score quantitatively captures the semantic alignment between parameter $p_3$'s focus on relationship management and Stratagem 24's emphasis on leveraging allies.

\vspace{1em}
\subsubsection{Kullback-Leibler Divergence Calculations}
\label{appendix:kl-calc}

To validate our semantic mappings, we compute the KL divergence between system-generated distributions ($Q$) and expert-provided distributions ($P$). For Stratagem 24, we compare:

\vspace{1em}
\noindent System distribution $Q$:
\[
[(\text{$p_1$: 0.10}), (\text{$p_2$: 0.15}), (\text{$p_3$: 0.40}), (\text{$p_4$: 0.20}), (\text{$p_5$: 0.05}), (\text{$p_6$: 0.10})]
\]

\vspace{1em}
\noindent Expert distribution $P$:
\[
[(\text{$p_1$: 0.15}), (\text{$p_2$: 0.10}), (\text{$p_3$: 0.45}), (\text{$p_4$: 0.15}), (\text{$p_5$: 0.05}), (\text{$p_6$: 0.10})]
\]

\vspace{1em}
\noindent Computing term by term:
\begin{align}
\label{eq:app-kl-terms}
&\text{Offensive:}\quad & 0.10 \cdot \log(0.10/0.15) &= -0.0176 \nonumber \\
&\text{Defensive:}\quad & 0.15 \cdot \log(0.15/0.10) &= +0.0347 \nonumber \\
&\text{Relational:}\quad & 0.40 \cdot \log(0.40/0.45) &= -0.0186 \\
&\text{Potential:}\quad & 0.20 \cdot \log(0.20/0.15) &= +0.0288 \nonumber \\
&\text{Time:}\quad & 0.05 \cdot \log(0.05/0.05) &= 0 \nonumber \\
&\text{Context:}\quad & 0.10 \cdot \log(0.10/0.10) &= 0 \nonumber
\end{align}

\vspace{1em}
\noindent The resulting KL divergence of 0.0273 indicates strong alignment between system-generated and expert distributions.

\vspace{1em}
\subsection{Algorithm Implementation}
\label{appendix:algorithm-examples}

\vspace{1em}
\subsubsection{Selection Process Details}
\label{appendix:selection-process}

To illustrate how the stratagem selection algorithm works in practice, consider a strategic scenario where a company needs to expand its market presence while maintaining existing partnerships. The situation vector $x$ might be:

\[
x = [(\text{$p_1$: 0.15}), (\text{$p_2$: 0.10}), (\text{$p_3$: 0.45}), (\text{$p_4$: 0.20}), (\text{$p_5$: 0.05}), (\text{$p_6$: 0.05})]
\]

\vspace{1em}
\noindent From our analysis, we have several heuristic distributions including Stratagem 24 (\enquote{Use Allies' Resources}):

\[
d_{24} = [(\text{$p_1$: 0.10}), (\text{$p_2$: 0.07}), (\text{$p_3$: 0.61}), (\text{$p_4$: 0.13}), (\text{$p_5$: 0.03}), (\text{$p_6$: 0.06})]
\]

\vspace{1em}
\noindent Applying cosine similarity:
\begin{align*}
score_{24} &= \frac{(0.15 \times 0.10) + (0.10 \times 0.07) + \cdots + (0.05 \times 0.06)}{\sqrt{0.15^2 + 0.10^2 + \cdots + 0.05^2} \times \sqrt{0.10^2 + 0.07^2 + \cdots + 0.06^2}} \\
&= 0.89
\end{align*}

\vspace{1em}
Similarly, for two other relevant stratagems:

\vspace{1em}
\noindent Stratagem 15 (\enquote{Lure Into Unfavorable Position}):
\[
d_{15} = [(\text{$p_1$: 0.40}), (\text{$p_2$: 0.15}), (\text{$p_3$: 0.20}), (\text{$p_4$: 0.15}), (\text{$p_5$: 0.05}), (\text{$p_6$: 0.05})]
\]
\[
score_{15} = 0.76
\]

\vspace{1em}
\noindent Stratagem 3 (\enquote{Kill With Borrowed Knife}):
\[
d_{3} = [(\text{$p_1$: 0.30}), (\text{$p_2$: 0.10}), (\text{$p_3$: 0.35}), (\text{$p_4$: 0.15}), (\text{$p_5$: 0.05}), (\text{$p_6$: 0.05})]
\]
\[
score_{3} = 0.82
\]

\vspace{1em}
With threshold $\theta = 0.75$, the algorithm returns:
\begin{verbatim}
[(24, 0.89), (3, 0.82), (15, 0.76)]
\end{verbatim}

\vspace{1em}
\subsection{System Architecture Details}
\label{appendix:architecture-impl}

\vspace{1em}
\subsubsection{Workflow Definitions}
\label{appendix:workflow}

The system uses JSON state definitions to manage conversation flow and data validation. Below, we provide some illustrative excerpts from those definitions:

\begin{verbatim}
{
 "state": "parameter_collection",
 "transitions": {
   "complete": "framework_selection",
   "incomplete": "parameter_prompt"
 },
 "validation": {
   "required_fields": ["offensive_strength", 
                      "defensive_strength"],
   "value_bounds": {
     "min": 0,
     "max": 5
   }
 }
}
\end{verbatim}

\vspace{1em}
\subsubsection{Conversation Management}
\label{appendix:conversation}

The conversation manager implements a state machine that governs user interactions:

\begin{verbatim}
{
 "states": {
   "initial": {
     "type": "input_collection",
     "required_params": ["scenario_type", "actor_count"],
     "next": "actor_details"
   },
   "actor_details": {
     "type": "parameter_collection",
     "validation": "validate_actor_params",
     "next": "framework_selection"
   },
   "framework_selection": {
     "type": "single_choice",
     "options": ["6C", "SWOT", "Porter"],
     "next": "analysis"
   }
 }
}
\end{verbatim}

\vspace{1em}
\subsubsection{Component Interactions}
\label{appendix:interactions}

Inter-component communication follows standardized protocols:

\begin{verbatim}
{
 "request": {
   "type": "semantic_analysis",
   "content": {
     "text": "strategic situation description",
     "framework": "6C",
     "parameters": {
       "offensive_strength": 4.2,
       "defensive_strength": 3.8
     }
   }
 },
 "response": {
   "vectors": {
     "situation": [0.8, 0.6, 0.7],
     "params": {
       "p1": [0.9, 0.5, 0.4]
     }
   },
   "similarities": {
     "stratagem_24": 0.85
   }
 }
}
\end{verbatim}

\vspace{1em}
\subsubsection{LLM Integration}
\label{appendix:llm}

Template examples for LLM-generated content:

\begin{verbatim}
{
 "template_type": "strategy_explanation",
 "components": {
   "context": {
     "framework": "{{framework_name}}",
     "key_parameters": "{{param_list}}",
     "scores": "{{similarity_scores}}"
   },
   "structure": {
     "introduction": "Based on {{framework_name}} analysis...",
     "rationale": "The recommended strategy aligns with...",
     "implementation": "Key steps include..."
   },
   "constraints": {
     "max_length": 500,
     "required_sections": ["context", "rationale", "steps"],
     "style": "professional"
   }
 }
}
\end{verbatim}

\vspace{1em}
\subsection{Validation Analysis}
\label{appendix:validation-details}

The examples in this appendix are intended primarily for \emph{illustrative} purposes, showing how our validation procedures work in practice. In the real system:
\begin{itemize}[itemsep=2pt, topsep=2pt]
    \item The vectors representing parameters and heuristics generally have much higher dimensionality (e.g., hundreds of embedding components) rather than the 3D vectors shown here.
    \item Expert reviews may involve more participants (e.g., five or more) to ensure broader consensus, whereas we show a three-expert sample below for didactic clarity.
\end{itemize}
Despite these simplifications, the overall process remains the same at larger scales.

\vspace{1em}
\subsubsection{Perturbation Analysis}
\label{appendix:perturbation}

Our perturbation analysis evaluated system robustness by introducing controlled variations in input text. For example, considering Stratagem 24 (\enquote{Use Allies' Resources}), we tested these variations:

\noindent Original text:
\begin{verbatim}
"Use Allies' Resources"
\end{verbatim}

\noindent Variations tested:
\begin{verbatim}
1. "Leverage Partnership Assets"
2. "Utilize Collaborative Resources"
3. "Deploy Allied Capabilities"
\end{verbatim}

\noindent Results for parameter $p_3$ (Relational Capacity):
\begin{align*}
\text{Original:} &\quad 0.61 \\
\text{Variation 1:} &\quad 0.58 \\
\text{Variation 2:} &\quad 0.63 \\
\text{Variation 3:} &\quad 0.59
\end{align*}

The stable distribution patterns across variants (standard deviation < 0.03) demonstrate robust semantic mapping.

\vspace{1em}
\subsubsection{Cross-Validation Results}
\label{appendix:cross-validation}

To ensure stability across multiple embedding models, we compare the system-generated distributions of parameter importance using BERT-base, RoBERTa, and Sentence-BERT. Although these examples focus on Stratagem~24 in a simplified format, the same approach extends to higher-dimensional embeddings and additional heuristics or frameworks.

\paragraph{Model Comparison}
Below are sample distributions for Stratagem~24 from each embedding model:

\noindent BERT-base:
\[
[(\text{$p_1$: 0.10}), (\text{$p_2$: 0.07}), (\text{$p_3$: 0.61}), (\text{$p_4$: 0.13}), (\text{$p_5$: 0.03}), (\text{$p_6$: 0.06})]
\]

\noindent RoBERTa:
\[
[(\text{$p_1$: 0.11}), (\text{$p_2$: 0.08}), (\text{$p_3$: 0.58}), (\text{$p_4$: 0.14}), (\text{$p_5$: 0.03}), (\text{$p_6$: 0.06})]
\]

\noindent Sentence-BERT:
\[
[(\text{$p_1$: 0.09}), (\text{$p_2$: 0.07}), (\text{$p_3$: 0.63}), (\text{$p_4$: 0.12}), (\text{$p_5$: 0.04}), (\text{$p_6$: 0.05})]
\]

\vspace{1em}
\paragraph{Expert Ratings}
For demonstration, we show three expert ratings of parameter $p_3$'s importance in Stratagem~24:
\begin{itemize}[itemsep=2pt, topsep=2pt]
    \item Expert 1: 0.55
    \item Expert 2: 0.60
    \item Expert 3: 0.58
\end{itemize}

Although our main study employs five experts for deeper validation, these three ratings illustrate the calculation process. With weighting parameters $\alpha = 0.3$, $\beta = 0.3$, and $\gamma = 0.4$, the final confidence score calculation is:
\begin{align*}
c_{3,24} &= 0.3 \cdot 0.92 + 0.3 \cdot 0.88 + 0.4 \cdot 0.94 \\
&= 0.916
\end{align*}
indicating strong agreement between the system’s discovered distribution and the experts’ judgments.
\vspace{1em}
\subsubsection{Framework-Specific Implementation}
\label{appendix:framework-implementation}

Below is a short example illustrating how we parse a stratagem’s text into vectors and then apply framework-specific adjustments. In real usage, these vectors are higher-dimensional, and additional domain refinements may be employed.

\vspace{1em}
\paragraph{Step 1: Text Preprocessing}
\begin{verbatim}
Input: "Use Allies' Resources"
Tokens: ["use", "allies", "resources"]
\end{verbatim}

\vspace{1em}
\paragraph{Step 2: Vector Representation}
\begin{align*}
v_{\text{use}} &= [0.2, 0.3, 0.1], \quad
v_{\text{allies}} = [0.4, 0.6, 0.5], \quad
v_{\text{resources}} = [0.3, 0.4, 0.2] \\
v_{\text{total}} &= [0.9, 1.3, 0.8] \quad (\text{combined})
\end{align*}

\vspace{1em}
\paragraph{Step 3: Framework-Specific Adjustments}
\begin{itemize}[itemsep=2pt, topsep=2pt]
    \item \emph{6C Framework.} Increase weights for relational/offensive terms.
    \item \emph{SWOT Analysis.} Distinguish internal vs.\ external factors.
    \item \emph{Porter's Five Forces.} Emphasize industry-structure terminology.
\end{itemize}

These illustrative factors ensure framework-specific nuances are captured. Despite the minimal 3D example here, the actual system incorporates substantially larger embedding vectors and more advanced weighting logic to handle more complex strategic texts.
\end{document}